\documentclass[letterpaper, 10 pt, conference]{ieeeconf}  %

\IEEEoverridecommandlockouts                              %

\usepackage{amsmath} %
\usepackage{algorithm}
\usepackage[noend]{algpseudocode}
\algrenewcommand\textproc{}
\usepackage{graphicx}

\usepackage{amssymb}
\usepackage{multirow}
\usepackage{xspace}
\usepackage{color}
\usepackage[normalem]{ulem}

\usepackage{array}
\usepackage{booktabs}
\makeatletter
\DeclareRobustCommand\onedot{\futurelet\@let@token\@onedot}
\def\@onedot{\ifx\@let@token.\else.\null\fi\xspace}
\def\eg{\emph{e.g}\onedot} 
\def\ie{\emph{i.e}\onedot}

\def\etal{\emph{et al}\onedot}
\makeatother

\def\etal{\emph{et al.}}
\def\ie{\emph{i.e. }}
\def\eg{\emph{e.g. }}

\newcommand{\real}{\mathbb{R}}

\def\xb{\mathbf{x}}
\def\yb{\mathbf{y}}
\newcommand{\thetab}{{\boldsymbol \theta}}

\def\msp{\hspace{5pt}}

\title{\LARGE \bf
	Optimal Solving of Constrained Path-Planning Problems with Graph Convolutional Networks and Optimized Tree Search
}

\author{
	Kevin Osanlou$^{1,2}$ 
	\qquad Andrei Bursuc$^3$
	\qquad Christophe Guettier$^1$
	\qquad Tristan Cazenave$^2$ 
	\qquad Eric Jacopin$^4$ 
	\\
	{\fontsize{11}{13}\selectfont$^1$Safran Electronics \& Defense \qquad  $^2$LAMSADE, Paris-Dauphine University \qquad $^3$valeo.ai}\\
	{\fontsize{11}{13}\selectfont$^4$CREC Saint-Cyr, Coetquidan School Campus}\\
	\tt\small{\{kevin.osanlou,christophe.guettier\}@safrangroup.com} \quad
	\tt\small{andrei.bursuc@valeo.com} \\
	\tt\small{tristan.cazenave@lamsade.dauphine.fr} \quad
	\tt\small{eric.jacopin@st-cyr.terre-net.defense.gouv.fr} \\
}

\begin{document}

	\maketitle

	\thispagestyle{empty}
	\pagestyle{empty}

	\begin{abstract}
		
		Deep learning-based methods are growing prominence for planning purposes. In this paper, we present a hybrid planner that combines a graph machine learning model and an optimal solver based on branch and bound tree search for path-planning tasks. More specifically, a graph neural network is used to assist the branch and bound algorithm in handling constraints associated with a desired solution path. There are multiple downstream practical applications, such as Autonomous Unmanned Ground Vehicles (AUGV), typically deployed in disaster relief or search and rescue operations. In off-road environments, AUGVs must dynamically optimize a source-destination path under various operational constraints, out of which several are difficult to predict in advance and need to be addressed online. We conduct experiments on realistic scenarios and show that graph neural network support enables substantial speedup and smoother scaling to harder path-planning problems. Additionally, information provided by the graph neural network enables the approach to outperform problem-specific handcrafted heuristics, highlighting the potential graph neural networks hold for path-planning tasks. 
		
	\end{abstract}

	\section{introduction}
Automated path-planning is an area of interest in AI with a wide panel of applications. The ability to efficiently plan an optimal path in a geometric graph that meets a set of requirements is becoming increasingly critical in a world where autonomy is starting to prevail. The requirements 
usually consist in a set of constraints imposed on the solution path 
, making it more difficult to compute. 
In the case of autonomous unmanned ground vehicles (AUGV), terrain structure is 
represented through a geometric graph, and maneuvers must consider terrain knowledge. Disaster relief, logistics, or area surveillance are a few among many applications for which online constrained path-planning algorithms enable autonomous mobility using perception and control functionalities. The ability of the AUGV to efficiently come up with an optimal path for a given mission has a direct impact on operational efficiency, 
underpinning the importance of an efficient path-planner. 

For such problems, classical robotic systems integrate A* algorithms \cite{Har68} as a best-first search approach in the space of available paths. For a complete overview of static algorithms (such as A*), replanning algorithms (D*), anytime algorithms (\eg ARA*), and anytime replanning
algorithms (AD*), we refer the reader to Ferguson \etal~\cite{Ferguson-2005-9201}. Tree search algorithms stemming from A* require the specification of a planning domain where constraints are modeled, and can be more or less efficient depending on the affinity of the search heuristic with the planning context. In this work, we focus on the branch and bound (B\&B) tree search algorithm instead \cite{Narendra1977}. More specifically, our study focuses on the performance gain when it is coupled with machine learning techniques.

\begin{figure}
\centering
\includegraphics[scale=0.15]{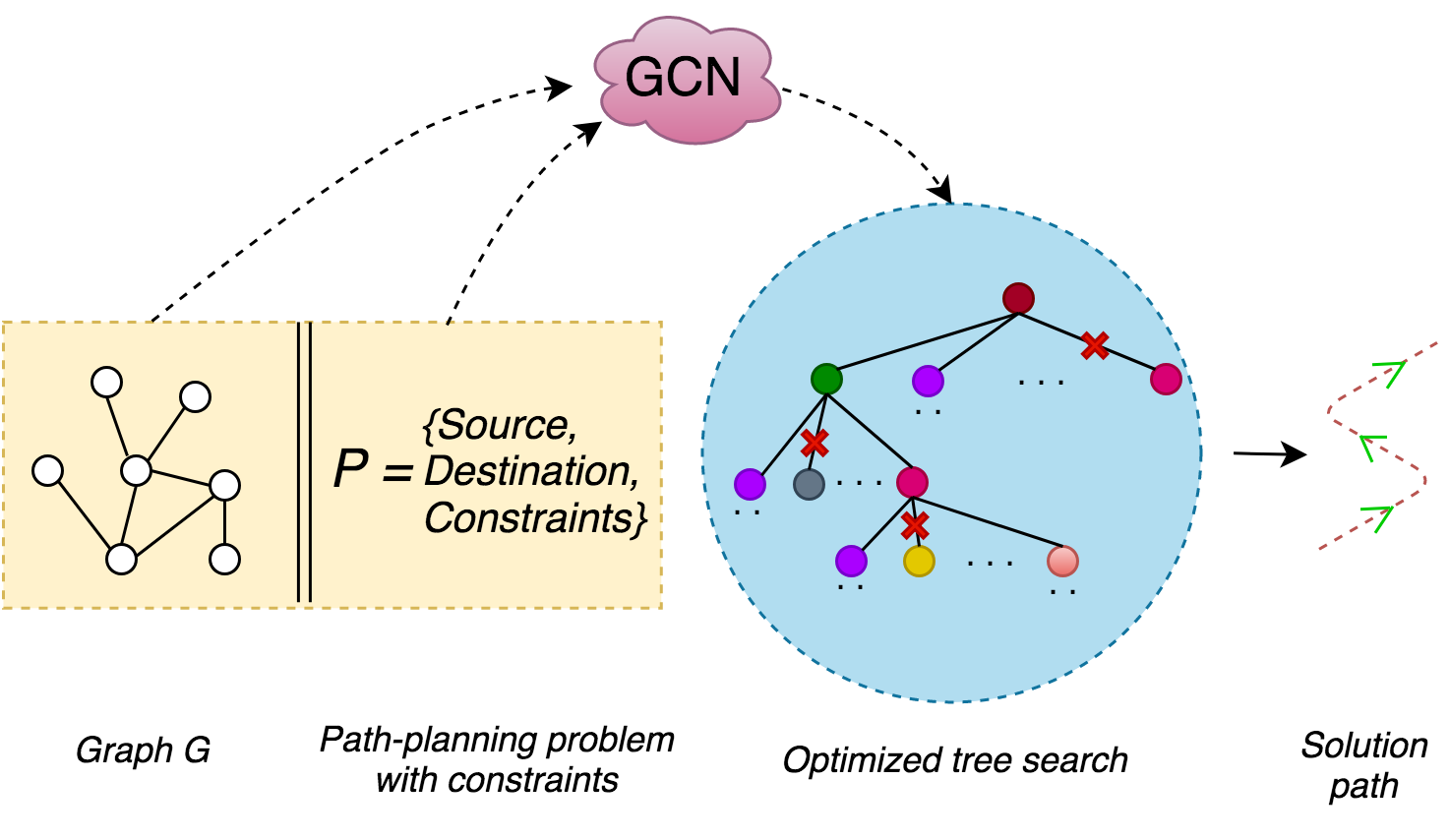}
\caption{Proposed framework for solving a path-planning problem with constraints in a graph with a GCN-assisted solver. The GCN takes as input the graph and the problem, and provides relevant information to speed up a tree search, after which an optimal solution path can be built.}
\label{fig_teaser}
\vspace{-6mm}
\end{figure}

Convolutional Neural Networks (CNNs) have been proven to be very efficient 
for computer vision applications, such as image recognition \cite{krizhevsky_2012}. They 
are a multilayer perceptron variant designed 
for minimal preprocessing and 
capable of detecting complex patterns in 
images.
In this paper, we are dealing with graphs which represent either maneuvers or off-road navigation. Instead of CNNs, the paper focuses on graph convolutional neural networks (GCNs), a recent 
architecture for learning complex patterns in graph data~\cite{Bruna2014,kipf_2017, YujiaLi2016}. 
One of the main reasons GCNs are preferred over CNNs for graph processing is that nodes and edges are attributed relevant features. CNNs are not able to learn features of an equivalent quality from an image of the graph. Moreover, CNNs are not invariant to node permutations, an issue GCNs do not share.

In this paper, we study a GCN-based approach for constrained path-planning and stick to exact resolution methods for path-related problems in a specific graph. We run experiments on realistic AUGV scenarios for which we consider mandatory pass-by nodes as type of constraint. This makes path-planning similar to the traveling salesman problem (TSP). The TSP is an NP-hard problem for which there exist efficient approximate solvers \cite{monnot2003approximation}, however it remains a challenge for exact approaches (\cite{caseau1997solving}, \cite{volgenant1982}, \cite{laporte1992traveling}). We make the following contributions. First, we define a GCN architecture suited for the considered problem. Second, we propose a self-supervised training strategy for the GCN. We then provide a framework which combines the GCN with the depth-first branch and bound (B\&B) algorithm. 
Finally, we 
conduct experiments on realistic problems for which results exhibit accelerated solving performance.
	\section{Related Work}
\label{sec:related}

In the past few years there has been a growing interest for transferring the intuitions and practices from neural networks on structured inputs 
towards graphs~\cite{gori_2005, henaff_2015, defferrard_2016, kipf_2017}.~\cite{henaff_2015} bridges spectral graph theory with multi-layer neural networks by learning smooth spectral multipliers of the graph Laplacian. Then ~\cite{defferrard_2016} and~\cite{kipf_2017} approximate these smooth filters in the spectral domain using polynomials of the graph Laplacian. Free parameters of the polynomials are learned by a neural network, avoiding the costly computation of the eigenvectors of the graph Laplacian.
We refer the reader to~\cite{bronstein_2017} for a comprehensive review on learning on graphs.

Applications of these types of networks are starting to emerge. Recent works suggest GCNs are capable of making key decisions to solve either path-planning problems in graphs~\cite{ZhuwenLi2018}, or even STRIPS planning tasks~\cite{Ma2018}. Regarding path-related optimization problems, Li\etal~\cite{ZhuwenLi2018} tackle the maximum independent set (MIS) problem with a solver that combines a GCN-guided tree search and local search optimization. The input to their solver is a graph, and the output is a highly optimized solution. Kool \etal~\cite{Kool2018} propose a reinforcement learning framework to solve the travelling salesman problem (TSP) and variants of the vehicle routing problem (VRP) approximately. While they prefer an encoder-decoder architecture over GCNs, they achieve better results than previous learning-based approaches. These related works focus on the approximate solving of a given task in big graphs. To this end, a learning model is coupled with tree search algorithms to narrow a wide search space in order to retrieve a high quality solution in a short time. In contrast, our work is intended for optimal task solving in smaller graphs. As optimal solving requires visiting most of the search space to ensure proof of optimality, previous approaches are not suitable and we proceed differently. 

	\section{Context and Problem Formalization}
\label{sec:context}
We consider a weighted connected graph $\mathcal{G}=(\mathcal{V},\mathcal{E},A)$, where $\mathcal{V}$ is the set of vertices, $\mathcal{E}$ the set of edges and $A$ the adjacency matrix of the graph. While in this work we choose to deal with realistic AUGV scenario graphs which are undirected (\S\ref{sec:experiments}), our approach is equally applicable to directed graphs.
In a typical crisis scenario, the AUGV has to proceed from an initial point to several areas for information gathering, before making its way to a final destination to share its assessment of the ongoing situation. Another frequent scenario consists in delivering food and first-aid equipment to areas likely to be undergoing a shortage. Such problems can be formalized mathematically.
Let $I$ be a path-planning problem instance, defined as follows: $$I = (start,dest,C)$$\begin{itemize}
\item $start \in \mathcal{V}$ is the index of the start node in $\mathcal{G}$,
\item $dest \in \mathcal{V}$ is the index of the destination node in $\mathcal{G}$,
\item $C$ is a set of constraints that need to be satisfied.
\end{itemize} 
Solving $I$ optimally means finding a path $p$, \ie a sequence of nodes (or edges), which begins from $start$, ends in $dest$, satisfies all constraints in $C$ and minimizes the total weight of the edges included in $p$. We can consider various types of constraints in $C$. In this work, we experiment with constraints related to mandatory nodes, which require the solution path to include a given set of nodes $M \in \mathcal{V}$. In the next sections, we will refer to a path-planning problem instance $(start,dest,M)$ simply as an instance. 
A valid solution path is required to include every node in $M$ at least once. Since the order of visit is not imposed for $M$, this problem can be assimilated to a TSP variant. %

	\section{Path-building with Graph Convolutional Networks}

In this section, we present our approach for training a graph neural network on a particular graph. 
We aim to leverage the learning capacity of the graph neural network to approximate the behavior of a model-based planner on the graph.

\subsection{Neural Networks}
Neural Networks (NNs) enable multiple levels of abstraction of data by using models with trainable parameters coupled with non-linear transformations of the input data.

In spite of the complex structure of a NN, the main mechanism is straightforward. A \emph{feedforward neural network}, or \emph{multi-layer perceptron (MLP)}, with $L$ layers describes a function $f(\xb; \thetab): \real^{d_{\xb}} \mapsto \real^{d_{\yb}}$ that maps an input vector $\xb \in \real^{d_{\xb}}$ to an output vector $\yb \in \real^{d_{\yb}}$. Vector $\xb$ is the input data that we need to analyze (\eg an image, a signal, a graph, etc.), while $\yb$ is the expected decision from the NN (\eg a class index, a heatmap, etc.). The function $f$ performs $L$ successive operations over the input $\xb$:
\begin{align}
h^{(l)} = f^{(l)}(h^{(l-1)}; \theta^{(l)}) = \sigma\left( \theta^{(l)} h^{(l-1)} + b^{(l)} \right)
\label{eq:layers}
\end{align}
where $h^{(l)}$ is the hidden state of the network and $f^{(l)}$ is the mapping function performed at layer $l$ and parameterized by trainable parameters $\theta^{(l)}$ and bias $b^{(l)}$, and piece-wise activation function $\sigma(\cdot)$; $h^{(0)}=\xb$. 

CNNs~\cite{fukushima_1982,lecun_1995} are a popular architecture for 2D data. They generalize MLPs by sliding groups of parameters across an input vector similarly to filters in image processing, leveraging fewer parameters and parallel computation. Hidden states in CNNs preserve the number of dimensions of the input, \ie 2D when images are used as input, and are called \emph{feature maps}.

\subsection{Graph Convolutional Networks}
GCNs are generalizations of CNNs to non-Euclidean graphs~\cite{bronstein_2017}. GCNs are in fact neural networks based on local operators on a graph $\mathcal{G}=(\mathcal{V},\mathcal{E})$ which are derived from spectral graph theory. The filter parameters are typically shared over all locations in the graph, thus the name \emph{convolutional}.

We consider here the approach of Kipf and Welling~\cite{kipf_2017}. The GCNs have the following layer propagation rule:
\begin{equation}
\textstyle
h^{(l+1)}= \sigma\!\left(\tilde{D}^{-\frac{1}{2}} \tilde{A}\tilde{D}^{-\frac{1}{2}}h^{(l)} \theta^{(l)} \right),
\label{eq:gcn-layer}
\end{equation}
where $\tilde{A} = A + I_{N}$ is the adjacency matrix of the graph with added self-connections such that when multiplying with $\tilde{A}$ we aggregate features vectors from both a node and its neighbors. Matrix  $I_N$ is the identity matrix; $\tilde{D}$ is the diagonal node degree matrix of $\tilde{A}$; $\sigma(\cdot)$ is the activation function, which we set to $\mathrm{ReLU}(\cdot) = \max(0,\cdot)$. Matrix $\tilde{D}$ is employed for normalization of $\tilde{A}$ in order to avoid a change of scales in the feature vectors when multiplying with $\tilde{A}$. In~\cite{kipf_2017}, the authors argue that using a symmetric normalization, \ie $\tilde{D}^{-\frac{1}{2}} \tilde{A}\tilde{D}^{-\frac{1}{2}}$, ensures better dynamics compared to simple averaging of neighboring nodes in the one-sided normalization $\tilde{D}^{-1}A$.

\subsection{Problem Instance Encoding}
The input of our model is a vector $\xb$ containing information about the problem instance, including the graph representation. For every instance $I = (start,dest,M)$ of a given graph $\mathcal{G}=(\mathcal{V},\mathcal{E})$, we associate a vector $\xb$ made up of triplet features from each node in $\mathcal{G}$, making up for a total of $3\times\left\vert{\mathcal{V}}\right\vert$ features. The three features for a node $j \in \mathcal{V}$ are:

\begin{itemize}
	
	\item \emph{start node} feature $st_{j}=1$ if node $j$ is the start node in instance $I$, otherwise $0$
	\item \emph{end node} feature $ed_{j}=1$ if node $j$ is the end node in instance $I$, otherwise $0$
	\item \emph{mandatory node} feature $my_{j}=1$ if node $j$ is a mandatory node in instance $I$, otherwise $0$.
\end{itemize}

\noindent
We obtain the input $\xb \in \real^{ \left\vert{\mathcal{V}}\right\vert \times 3}$ by stacking node features: 
\begin{equation}
\xb= \lbrack\lbrack st_{1},ed_{1},my_{1}\rbrack,\lbrack st_{2},ed_{2},my_{2}\rbrack, ..., \lbrack st_{\left\vert{\mathcal{V}}\right\vert},ed_{\left\vert{\mathcal{V}}\right\vert},my_{\left\vert{\mathcal{V}}\right\vert}\rbrack\rbrack 
\label{equ:embedding}
\end{equation}

\subsection{Neural network architecture and training}
We define a neural network $f$ that consists of a sequence of multiple graph convolutions followed by a fully connected layer. This GCN takes as input any instance $I$ over the graph $\mathcal{G}$, and outputs a probability vector $\hat{\yb} \in \real^{\left\vert{\mathcal{V}}\right\vert}$. Then, $\arg\max(\hat{\yb})$ corresponds to the next mandatory node to visit from the $start$ node in an optimal path that solves $I$. 
The hidden states of the graph convolution layers $h^{(l)}\in \real^ {{\left\vert{\mathcal{V}}\right\vert} \times N_{hidden}}$ consist of $N_{hidden}$ higher-dimensional features  for each node in the graph . After the final convolutional layer, the feature matrix
is flattened into a vector by concatenating its rows and linked to a fully connected layer that maps it to a vector $z \in \real^{\left\vert{\mathcal{V}}\right\vert}$. We use the \emph{softmax} function to convert $z$ into probabilities.

NNs trained in a supervised manner use labeled training data, \ie a set of input-output pairs $(\xb_i, \yb_i)$ sampled from a large training set. Here, $\xb_i$ is an instance $I$ and $\yb_i$ is the index of the next mandatory node for $\xb_i$ in an optimal path solving $I$. We train the GCN on instances that have already been optimally solved by an exact planner, which serves here as a \emph{teacher}. The network learns to approximate the solutions 
computed by the planner. To this end, we train the network using the \emph{negative log-likelihood loss}
, used for multi-class classification, and stochastic gradient descent (SGD).

\subsection{Mandatory Node Ordering}
\label{sec:order}
For a given input instance, the GCN computes a mandatory node prediction at a time. In order to make it compatible with instances with varying amounts of mandatory nodes we make a few adjustments. Given an instance with multiple mandatory nodes $(start, dest, M)$, we perform multiple recursive GCN calls to get the next mandatory node predictions $q_i$. More specifically, after a prediction $q_i$ is computed, we generate a sub-instance where the start node becomes the current predicted mandatory node $q_i$ and where the list of mandatory nodes contains the remaining nodes except $q_i$, \ie $(q_i, dest, \tilde{M}_i=\tilde{M}_{i-1} \setminus q_i)$, where $\tilde{M}_0=M$ and $q_0=start$. We use the recursive calls during both training and testing. An interesting side-effect of this strategy is that it ensures an implicit balancing of the training samples by difficulty, as we generate sub-instances ranging from challenging (large $\lvert M \rvert$) to trivial (small $\lvert M \rvert$).

	\section{Self-supervised learning}
\label{sec:data}
We present a self-supervised learning strategy aimed at training the GCN for a particular graph. First, we define a planning domain to solve instances. Second, we introduce a modified version of the A* algorithm for generating optimal data on which the GCN is trained. In this section, we refer to an instance as a planning state $s = (start,dest,M)$. An end state is a termination instance, \ie an instance for which the destination node has been reached and all mandatory nodes have been visited. We denote the termination instances as $F_i = (i,i,\emptyset), i \in \{1, 2, ..., \left\vert{V}\right\vert\}$. There are exactly as many termination instances as there are nodes in $\mathcal{G}$. We respectively define the \emph{successors} and \emph{predecessors} of a state $s = (start,dest,M)$ as $succ(s)$ and $pred(s)$ in Table~\ref{tab:table01}.

\begin{table}[ht]
\caption{Transition rules to successors and predecessors.}
\vspace{-5mm}
\begin{center}
\small
\begin{tabular}{@{\msp}l@{\msp} | @{\msp}l@{\msp}}
\toprule
\multicolumn{2}{c@{\msp}}{\textbf{$s = (start,dest,M)$} } \\\midrule
\multicolumn{1}{c|@{\msp}}{Successor state $s'$} &  \multicolumn{1}{c@{\msp}}{Predecessor state $s'$}\\
\midrule
   $s' = (start',dest',M')$ &  $s' = (start',dest',M')$\\
   $(start,start') \in \mathcal{E}$ & $(start',start) \in \mathcal{E}$ \\
   $dest' = dest$ & $dest' = dest$  \\
   $M' = M \backslash \{start'\}$ & $M' = M $ \Comment{$^1*$} \\
   \multicolumn{1}{c|@{\msp}}{-} & or \\
   \multicolumn{1}{c|@{\msp}}{-} & $M'=M\cup\{start\}$ \Comment{$^2*$} \\
    $s$ is not an end state & $s'$ is not an end state  \\
    Transition cost: $(start,start')$ & Transition cost: $(start',start)$ \\
\bottomrule
\end{tabular}
\label{tab:table01}
\end{center}
 \footnotemark{$*$ only if $start' \not \in M$\\}
 \footnotemark{$*$ only if $start \neq dest$}
\end{table}

Transition costs from a state to a neighboring state is the cost of the edge in the graph linking the start nodes of both states. With these rules, the destination node remains always the same. Therefore, we run the backwards version of A* from every termination instance $F_i$ as initial state (using $pred$ as rule of succession). For each state $s$ visited by A*, a path $p$ is built from $s$ to $F_i$ which the algorithm considers the shortest. We define $g(s)$ as the cost of $p$, $a(s)$ as the next state visited after $s$ in $p$, and $d(s)$ as the first mandatory node of $s$ that is visited in $p$. 

Furthermore we perform the following changes to A*. First, when a shorter path is found to a state $s'$ while developing a state $s$, \ie $g(s') > g(s) + c(s, s')$, the values of $a(s')$ and $d(s')$ are also updated along with $g(s')$ to take into account the shorter path. Secondly, we set the heuristic function $h$ to $0$. Since A* is run backwards from a termination state $F_i$, we are not aiming for the algorithm to reach a defined state in particular, but seek to reach as many states as possible. This ensures that when a state $s$ is taken from the OPEN priority list of states left to develop, an optimal path from $s$ to $F_i$ has already been found. Choosing $d(s)$ as the next mandatory node to visit thus enables optimal solving. Consequently we can add the pair $\langle s,d(s)\rangle$ to the training set. We provide the pseudo-code in Algorithm~\ref{a_star}.

\begin{algorithm}[tb]
\small
\caption{Backwards A* for reverse instance solving}\label{a_star}
\begin{algorithmic}[1]
\Function{ComputePaths()}{}
	\While{$elapsed$ $time < timeout$}
		\State remove state $s$ from the front of OPEN;
		\If{$length(mand(s)) > 0$} \Comment{$^3*$}
			\State insert the pair $<s,d(s)>$ to the training set 
		\EndIf 
		\ForAll{$s' \in pred(s)$} \Comment{$^3*$}
		 \If {$g(s') > g(s) + c(s, s')$}
		 	\State g(s') = g(s) + c(s, s')
		 	\State a(s') = s
		 	\If {$mand(s) = mand(s')$}
		 	\State $d(s') = d(s)$ \Comment{$^4*$}
		 	\Else 
		 	\State $d(s') = startNode(s)$ \Comment{$^{3,5}*$}
		 	\EndIf
		 	\State insert $s'$ into OPEN with value $g(s')$\Comment{$^6*$}
		 \EndIf
		\EndFor
	\EndWhile

\EndFunction
\Function{main()}{}
	\ForAll{$i \in \{1, 2, ..., \left\vert{V}\right\vert\}$}
		\ForAll{$s \in S$}
			\State $g(s) = \infty$
		\EndFor
		\State $g(E_i) = 0$, $a(E_i) = \emptyset$, $d(E_i) = \emptyset$
		\State $OPEN = \emptyset$
		\State insert $F_i$ into OPEN with value $g(E_i)$ \Comment{$^6*$}
		\State ComputePaths();
	\EndFor
\EndFunction
\end{algorithmic}
\footnotemark{$*$ $length(x)$: returns length of list $x$; $mand(x)$: returns mandatory nodes of instance $x$; $startNode(x)$: returns the start node of instance $x$; $pred(x)$: returns all possible predecessors of instance $x$\\}
\footnotemark{\footnotesize{$*$ $s'$ is created from $s$ without adding its start node to the list of mandatory nodes}\\}
\footnotemark{\footnotesize{$*$ $s'$ is created from $s$ by adding its start node to the list of mandatory nodes}\\}
\footnotemark{\footnotesize{$*$ $h = 0$}}

\end{algorithm}

The data generated by the algorithm is added to the training set and shuffled. The GCN is then trained on this set with supervised learning. Results show that training on the "synthetic" data generated with A* enables the GCN to generalize well on instances that A* did not process. We argue this is because the distribution obtained with A* is related to the path length of resolved instances. In fact, graph patterns already explored in short path solutions are incrementally included into longest ones.

	\section{Depth-First Branch and Bound Tree Search}
\label{sec:branch}
It is possible to resolve an instance within a combinatorial search tree, rather than in the path-planning domain defined in \S\ref{sec:data}. To this end, we compute the shortest source-destination paths for every pair of nodes $\langle i,j \rangle$ in $\mathcal{G}$ using Dijkstra's algorithm, as well as the associated path cost. Solving an instance $I= (start, dest, M)$ then becomes equivalent to finding the optimal order in which the mandatory nodes in $M$ are visited for the first time. The solution path associated with an order $o = (m_1, m_2, m_3, ... , m_{q-1}, m_q)$ of the mandatory nodes can be built by concatenating the shortest path from $start$ to $m_1$, from $m_1$ to $m_2$, from $m_2$ to $m_3$, ... , from $m_{q-1}$ to $m_q$, and from $m_q$ to $dest$. Its total cost is the sum of the cost of each shortest path used to build it. A particular tree can be searched for identifying an optimal order of the mandatory nodes. For an instance $I$, we define the root of this tree as the $start$ node, every leaf node as the $dest$ node, and every intermediate tree node as a mandatory node in $M$, such that a path from the root of the tree to a leaf defines an order in which to visit the mandatory nodes in $M$. The cost of of transitioning from a node $v$ to a child node $v'$ in the tree is the cost of the shortest path in $\mathcal{G}$ between the pair $\langle v,v' \rangle$. In the following, we refer to this tree as the mandatory search tree. Since we are using here
only mandatory node constraints, the combinatorial optimization problem associated with this tree displays a similar structure with the TSP. However, it differs in the choice of the start and destination nodes which are fixed. %

The branch and bound algorithm (B\&B) is a popular tree search algorithm that is well known for its computational efficiency (\cite{Narendra1977}). In this work, we consider a depth-first B\&B search algorithm. When developing a node inside the tree, the algorithm checks if each branch is expected to host a better solution than the best solution found so far. Should that not be the case for a given branch, the branch is cut, and the algorithm will not develop nodes further down the branch. This is done by using a lower bound and an upper bound. The lower bound is the sum of the total cost from the root node to the current node and a heuristic function $\pi$
that approximates the remaining cost from the current node to the best achievable solution in branches below. This heuristic function should return a value as close as possible to this remaining cost (to cut as frequently as possible), while staying smaller (for the algorithm to remain optimal).

Next, we define the heuristic function $\pi$ that we use. Let $v \in M$ be a mandatory node in the mandatory search tree, and $R$ the set of remaining mandatory nodes left to the leaf node $dest$, \ie the nodes in $M$ that haven't been included between the root node $start$ and $v$.  Let $D = \{v\} \cup R \cup \{dest\}$. We define two functions, $min_1$ and $min_2$, that respectively return for a node $x$ in $D$ the lowest shortest path cost in $\mathcal{G}$ from $x$ to any node in $D \backslash \{x\}$, and the second lowest such cost. We build the heuristic function $\pi$ by considering the remaining nodes left. For each node left, we consider the weight of all edges connecting it to other nodes in $D$ and add the first and second smallest such weights, with the exception of $v$ and $dest$ for which only the first smallest weight is added, and divide the total by 2: 

$$\pi(v) = \frac{1}{2}(min_1(v) + min_1(dest) + \sum_{r \in R}min_1(r)+min_2(r))$$

\begin{figure}
\centering
\includegraphics[scale=0.075]{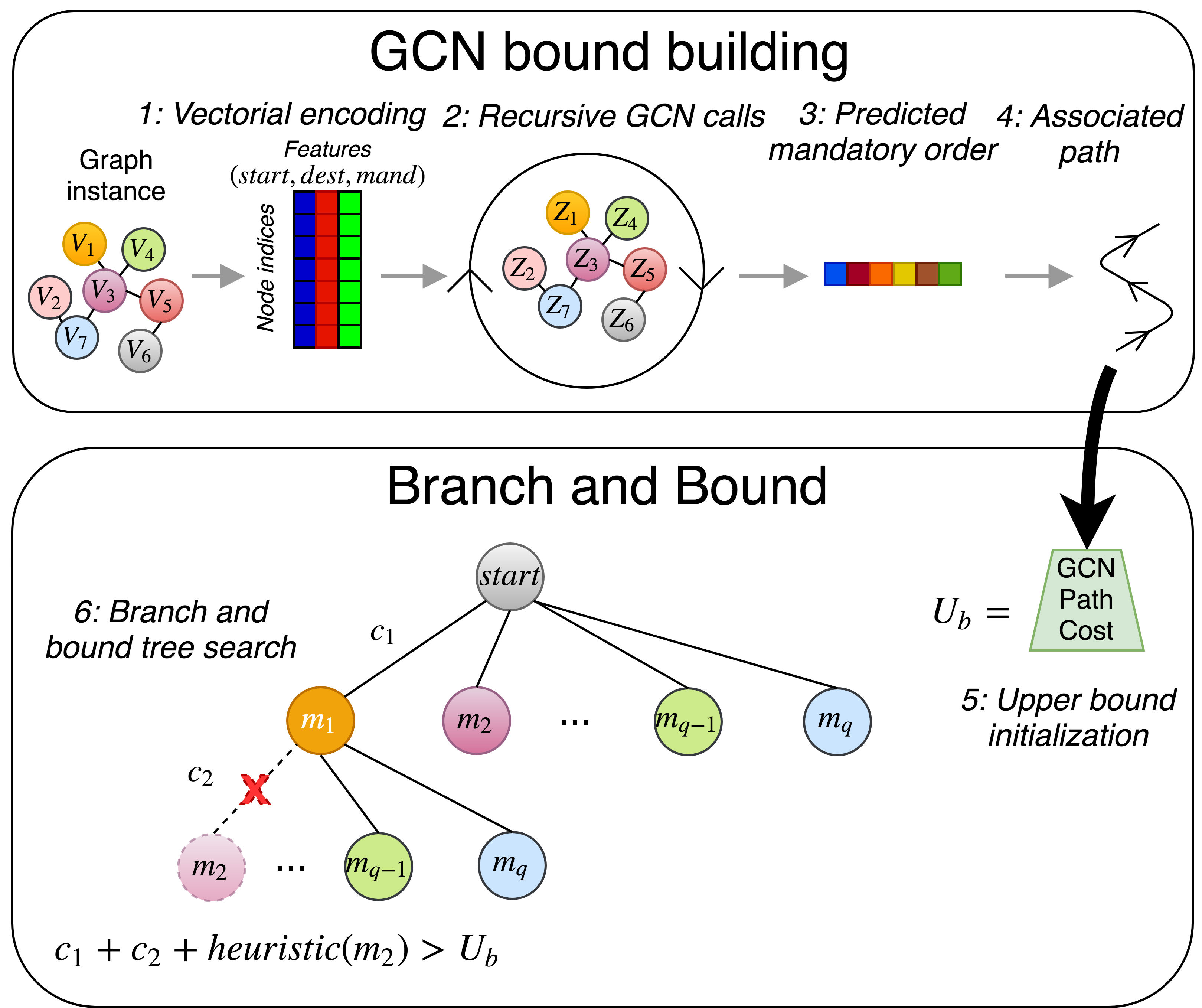}
\caption{GCN-assisted branch and bound algorithm pipeline. The GCN is used recursively to build an order of visit for the mandatory nodes. The order is converted into a path using the previously computed shortest path pairs, and the cost of the path is used as an initial upper bound for the algorithm. Here, the upper bound of the GCN allows for a level 2 early cut.}
\label{fig_bnb_ub_cut}
\vspace{-3mm}
\end{figure}

Recent progress in learning-assisted tree search has shown that machine learning can be used to narrow the search space in very large domains to allow for efficient solving. Inspiring results have been shown in the game of Go
~\cite{silver_2016, silver_2017}, where a very good solution, which is not required to be optimal, is found in record time. Promising parts of the search tree are first visited in accordance with the suggestions of the neural network, unpromising parts then, only if time allows. On the other hand, in a context where finding an optimal solution is critical, the search cannot be directed in such a way, as proof of optimality is required. Consequently, we keep the GCN out of the tree search procedure. For a given instance, we use the GCN recursively in order to obtain a suggested order of visit of the mandatory nodes (\ref{sec:order}), from which we build the associated solution path by concatenating the shortest paths. This is done in negligible time compared to the tree search. The cost of the solution path found in such {\it probing} manner \cite{Osanlou2018ConstrainedSP} is then used as an initial upper bound for the B\&B algorithm (figure \ref{fig_bnb_ub_cut}). In section \S\ref{sec:experiments} we
conduct experiments to evaluate the influence of the upper bound obtained with the GCN on search performance.

	\section{Experiments}
\label{sec:experiments}
\subsection{Benchmarks and baselines}
We run experiments to evaluate the impact of the GCN's upper bound on the B\&B method. Since proof of optimality is necessary in our context, we focus only on small-scale problems for which optimal solving is possible in reasonable time. We consider four different graphs, $\mathcal{G}_1$, $\mathcal{G}_2$, $\mathcal{G}_3$ and $\mathcal{G}_4$, with respectively 15, 23, 22 and 23 nodes. These graphs represent realistic AUGV crisis scenarios in which aid has to be provided to key points in operational areas. More details on the graphs are available in \cite{gue07}, from which the scenarios have been built. We generate $1512$, $2928$, $2712$  and $2712$ random instances for each graph respectively. In order to remain close to some 'realistic' instances, we generate the instances as follows: using the shortest source-destination paths computed previously, we apply a decimation ratio (typically 80\%) to keep only the 20\% source-destination pairs that have the longest shortest paths. For each resulting pair $\langle start, dest \rangle$ kept, we generate multiple random instances with an increasing cardinality for the set of mandatory nodes, ranging from 5 to 12. 

We consider 4 different baseline solvers on the benchmark instances generated to compare solving performance. All solvers search for an optimal solution path. First, we use a solver based on dynamic programming (DP) which searches the mandatory search tree. Second, we run the B\&B algorithm to search the mandatory search tree, both with and without the upper bound provided by the GCN. Lastly, we solve instances using forward A* applied on the planning domain described in section \S\ref{sec:data}, with the minimum spanning tree (MST) as heuristic function. The MST heuristic is computed for an instance ($start$, $dest$, $M$) by considering the complete graph $\mathcal{G}'$, which comprises only the $start$, $dest$ and mandatory nodes $M$. All pairs of nodes $(v,v')$ in $\mathcal{G}'$ are connected by an edge which has a weight equal to the cost of the shortest path from $v$ to $v'$ in $\mathcal{G}$. The MST heuristic value is obtained by adding the following three values: the total weight of the MST of all mandatory nodes $M$ in $\mathcal{G}'$, the minimum edge weight in $\mathcal{G}'$ from the $start$ node to any node in the MST, and the minimum edge weight in $\mathcal{G}'$ from any node in the MST to the $dest$ node. 

\subsection{Implementation details}
We set the run-time of A* to 10 hours per graph for data generation. We use 3 graph convolutional layers of width 100. During training, we apply batch normalization \cite{ioffe_2015} with decay of moving average $\varepsilon = 0.9$, dropout with \emph{drop rate} of $0.1$, and train the GCN with \textit{Adam} \cite{diederik-kingma:adam}. We set the learning rate to $\eta=10^{-4}$. We train models on a Tesla P100 GPU using over 1.5M training examples generated by A*, for $\sim$5 hours. We conduct the benchmark tests in this section on a laptop with an Intel i5 processor and 8GB of RAM. We point out that our approach requires GPU only for the training of the GCN, which can be done offline. Problem instances can be then solved online on a CPU.

\begin{figure}[tb]
\centering
\includegraphics[scale=0.7]{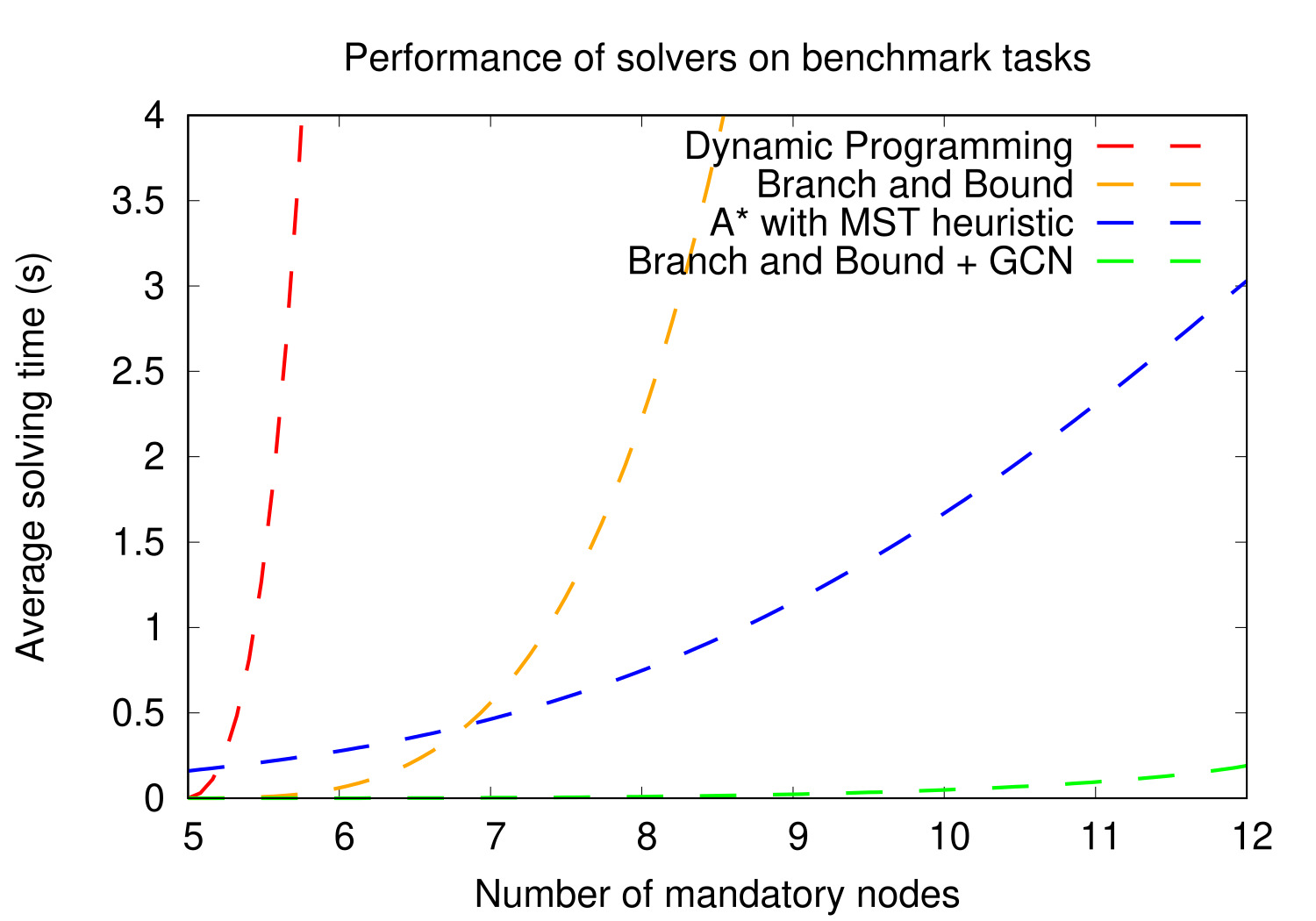}
\caption{Comparison of the performance of different solvers on benchmark instances generated for graph $\mathcal{G}_2$. The $X$ axis represents the number of mandatory nodes of the instances, the $Y$ axis the average solving time. We limit the $Y$ axis to a range [0,4] to obtain a better comparison scale.}
\label{curve}
\vspace{-3mm}
\end{figure}

\begin{table}[b]
\begin{center}
\caption{Experiments for graph $\mathcal{G}_3$. Legend: T/O = $5$ minutes timeout.}
\label{tab:table02}
\footnotesize
\begin{tabular}{ @{\msp}l@{\msp} | @{\msp}c@{\msp}@{\msp}c@{\msp}@{\msp}c@{\msp}@{\msp}c@{\msp}@{\msp}c@{\msp}}
\toprule
Mandatory \#: & 5 & 7& 9 & 11 & 12\\
\midrule
DP & & & & & \\
Avg. node visits:& 11.7K & 876K & 98.6M & T/O & T/O\\
Avg. time (s):& - & 0.26 & 30.52 & T/O & T/O\\
\midrule 
B\&B & & & & & \\
Avg. node visits:&418 & 4,94K & 146K & 11,1M & 120M\\
Avg. time (s):& - & - & 0.07 & 5.39 & 71.1\\
\midrule
A*, h=MST & & & & & \\
Avg. state visits:& 26.4 & 58.5 & 141 & 342& 503\\
Avg. time (s):& 0.14 & 0.29 & 0.73 & 1.84 & 2.70\\
\midrule  
B\&B + GCN & & & & & \\
Avg. node visits:& 148 & 1,24K & 10,8K & 161K& 642K\\
Avg. time (s):& - & - & 0.01 & 0.08& 0.34\\ 
\bottomrule
\end{tabular}
\end{center}
\end{table}

\begin{table}[htb]
\begin{center}
\caption{Experiments for graph $\mathcal{G}_4$. Legend: T/O = $5$ minutes timeout.}
\label{tab:table03}
\footnotesize
\begin{tabular}{ @{\msp}l@{\msp} | @{\msp}c@{\msp}@{\msp}c@{\msp}@{\msp}c@{\msp}@{\msp}c@{\msp}@{\msp}c@{\msp}}
\toprule
Mandatory \#: & 5 & 7& 9 & 11 & 12\\
\midrule
DP & & & & & \\
Avg. node visits:& 11.7K & 876K & 98.6M & T/O & T/O\\
Avg. time (s):& - & 0.24 & 30.53 & T/O & T/O\\
\midrule
B\&B & & & & & \\
Avg. node visits:& 545 & 7.58K & 193K & 11,8M & 122M\\
Avg. time (s):& - & - & 0.09 & 5.28 & 70.8\\
\midrule
A*, h=MST & & & & & \\
Avg. state visits:& 41.78 & 96.8 & 226 & 787& 1,09K\\
Avg. time (s):& 0.15 & 0.44 & 0.96 & 3.18 & 4.36\\
\midrule  
B\&B + GCN & & & & & \\
Avg. node visits:& 233 & 2.46K & 29.1K & 235K& 859K\\
Avg. time (s):& - & - & 0.02 & 0.11& 0.56\\ 
\bottomrule
\end{tabular}
\end{center}
\vspace{-5mm}
\end{table}

 \subsection{Results}
We summarize results for $\mathcal{G}_2$ in Figure \ref{curve}. We detail the experiments for graph $\mathcal{G}_3$ and $\mathcal{G}_4$ in Tables \ref{tab:table02} and \ref{tab:table03}. We include solving time only if it is measurable by CPU clock time. Figures for all graphs show a similar trend. We note that for instances with seven mandatory nodes and more, best-first algorithms such as A* applied on the planning domain defined in \S\ref{sec:data} become more suited than depth-first algorithms such as B\&B applied on the combinatorial search tree associated with the mandatory constraints. 
Although each planning state takes longer to compute in order to account for the specifics of the planning domain, overall significantly fewer planning states are visited than mandatory search tree nodes. This is because the planning domain takes advantage of the graph structure, which gives A* a significant edge over depth-first DP and B\&B. 
We note, however, that when the upper bound of the GCN is used, the B\&B algorithm is able to outperform A* on all instances, even the most complex ones, while scaling more smoothly with the number of mandatory nodes. 
In table \ref{tab:table04} we provide additional insight on these results through information collected from the mandatory search tree for instances with 11 mandatory nodes.
The average number of nodes processed in the subtree of each child node of the root node is given, as well as the average score of the best known solution after the subtree is processed.

\begin{table}[htb]
\begin{center}
\caption{Information from the mandatory search tree.}
\label{tab:table04}
\footnotesize
\begin{tabular}{ @{\msp}l@{\msp} | @{\msp}c@{\msp}@{\msp}c@{\msp}@{\msp}c@{\msp}@{\msp}c@{\msp}@{\msp}c@{\msp}}
\toprule
 Root node child \# & - & 1 & 2 & 3 & ...\\
\midrule
B\&B & & & & & \\
    Avg. node visits& - & 11M & 50K & 30K & ...  \\
    Avg. best sol. score& $\infty$ & 10.35K & 9940 & 9781 & ... \\
\midrule 
B\&B + GCN & & & & & \\
Avg. node visits& - & 74K & 11K & 7K & ...\\
Avg. best sol. score& 9890 & 9300 & 9264 & 9249 & ...\\ 
\bottomrule
\end{tabular}
\end{center}
\vspace{-2mm}
\end{table}

Since our B\&B algorithm is depth-first,  processing the entire subtree under the first child node of the root node when no initial upper bound is known is highly computationally expensive. Indeed, no cut can be made until a leaf node is reached, and even then, the identified solution is very likely to be costly compared to the optimal solution, thus the updated upper bound would still not allow for frequent cuts, until a good part of the subtree has been processed. On the other hand, if a good upper bound is known in advance, which is generally the case for the one given by the GCN in our experiments, the algorithm does not suffer from this issue, and early cuts can be made.

	\section{Discussion and Further Works}

We experiment with path-planning problems defined by three features: the start node, destination node, and mandatory nodes. 
We accelerate optimal depth-first solving of the search tree associated with the mandatory constraints by leveraging the upper bound computed by the GCN. We show that this speedup is significant, competing successfully with A*.
This is the case even for scenarios where handling constraints within the planning domain is more appropriate than extracting and solving them separately.
Also, our attempts to guide A* search with a GCN heuristic achieved worse results than the MST heuristic. The reason is due to the best-first approach for which the GCN is unable to provide a suitable heuristic. 
Moreover, the proposed framework can include additional types of constraints for path-planning problems. Each new constraint type results in additional features on nodes, and potentially also on edges \cite{vos2017}. In this case, the GCN would learn to predict the next node to visit, and not the next mandatory node. Recursive GCN calls would be made until a solution which satisfies all constraints is found, backtracking when necessary. The approach can be combined with state-of-the-art constraint propagation techniques \cite{guettier_lucas_2016}. In the same manner, the solution cost can be used as an initial upper bound for a depth-first search of a combinatorial tree associated with the constraints. 

Learning-wise, the more constraint types there are for a path-planning problem, the wider the GCN learning domain will be. Further work will especially focus on this limitation to relate the exhaustiveness of the training phase with the variety of constraint types. Also, the proposed approach requires neural network offline training for a given graph (\eg problem scenario). It can then be used online for path re-planning purposes, as the AUGV drives through the graph, with 
appealing computational performances.

	\section{Conclusion}
	
	We introduced a method combining graph neural networks and branch and bound (B\&B) tree search to handle constraints in path-planning, successfully accelerating optimal solving of path-planning tasks. A relevant self-supervised strategy has been developed, based on A*, which provides appropriate data to train the graph neural network. The heuristic information computed by the graph neural network enables better scaling of the B\&B algorithm onto more complex problems. Results exhibit solving times that outperform A* with problem-specific handcrafted heuristics. Various path-planning applications for autonomous vehicles can benefit from such an approach, especially when known terrains are given and path or itineraries must be computed on the fly. We also hope this line of work will serve to highlight the merits of using graph neural networks for path-planning tasks.
	

	\nocite{osanlou2021learning}
	\nocite{li2021training}	
	\nocite{zhao2021distributed}
	\nocite{park2021learning}	
	\nocite{zhang2020learning}
	\nocite{wang2020learning}	
	\nocite{li2020graph}
	\nocite{zhou2020variational}	
	\nocite{silver2020planning}
	\nocite{hameed2020reinforcement}	
	\nocite{zhou2020reinforced}
	\nocite{cappart2021combinatorial}	
	\nocite{sato2019approximation}
	\nocite{prates2019learning}	
	\nocite{hu2020petri}
	\nocite{li2021message}	
	\nocite{rusek2018message}
	\nocite{hu2020collaborative}	
	\nocite{gama2021graph}
	\nocite{hu2020reinforcement}	
	\nocite{weng2020joint}
	\nocite{lemos2019graph}	
	\nocite{drori2020learning}
	\nocite{vesselinova2020learning}	
	\nocite{Guo2019SolvingCP}
	\nocite{Wu2019ACS}	
	\nocite{Xu2019HowPA}
	\nocite{Liu2020TowardsDG}	
	\nocite{Ying2019GNNExplainerGE}
	\nocite{Wang2019DeepGL}
	\nocite{Rossi2020SIGNSI}	
	\nocite{Garg2020GeneralizationAR}	
	\nocite{Loukas2020WhatGN}
	\nocite{osanlou2021learning}
    \nocite{lin2020goal}
    \nocite{lin2020gaussian}    
	\nocite{sombolestan2019optimal}
	\nocite{otte2015survey}
	\nocite{zohdi2020game}
	\nocite{yonetani2021path}
	\nocite{sung2021training}
	\nocite{xin2005neural}
	\nocite{zhou2021ast}
	\nocite{rusek2020routenet}
	\nocite{zhang2017type}
	\nocite{glasius1995neural}
	\nocite{wu2020learning}
	\nocite{li2021hierarchical}
	\nocite{chen2021graph}
	\nocite{cao2021spectral}
	\nocite{syed2014guided}
	\nocite{li2021interactive}
	\nocite{lu2021mgrl}
	\nocite{rehder2017driving}
	\nocite{peng2019self}
	\nocite{casas2020spagnn}
	\nocite{fu2020magnn}
	\nocite{lu2020lstm}
	\nocite{wai2019adaptive}
	\nocite{weng2021ptp}
	\nocite{chen2021graph}
	\nocite{monti2021dag}
	\nocite{lee2019joint}
	\nocite{wiederer2020traffic}
	\nocite{tolstaya2020learning}
	\nocite{wang2020neural}
	\nocite{low2019solving}
	\nocite{yan2020towards}
	\nocite{guo2020autonomous}
	\nocite{qie2019joint}
	\nocite{qu2020novel}
	\nocite{yao2020path}
	\nocite{bae2019multi}
	\nocite{lei2018dynamic}
	\nocite{lakshmanan2020complete}
	\nocite{wang2020mobile}
	\nocite{bency2019neural}
	\nocite{qureshi2018deeply}
	\nocite{li2018neural}
	\nocite{saraswathi2018optimal}
	\nocite{ichter2020learned}
	\nocite{xing2020graph}
	\nocite{peng2019self}
	\nocite{fu2018improved}
	\nocite{qureshi2020motion}
	\nocite{chen2020relational}
	\nocite{aggarwal2020path}
	\nocite{madridano2021trajectory}
	\nocite{prianto2020path}

	\bibliography{bnb}
	\bibliographystyle{IEEEtran}

\end{document}